\documentclass[pdflatex,sn-nature]{sn-jnl}
\usepackage{graphicx}
\usepackage{multirow}
\usepackage{amsmath,amssymb}
\usepackage{booktabs}
\usepackage{array}
\usepackage{url}
\usepackage{xcolor}
\usepackage{placeins}
\begin{document}

\title[AI-inferred expressed well-being and collective-action discourse]{AI-inferred expressed well-being and collective-action discourse}

\author*{\fnm{Wentao} \sur{Xu} \textnormal{(Doctor of Informatics)}}\email{myrainbowandsky@gmail.com}

\abstract{Climate campaigns are often evaluated through attention and mobilization, but less is known about the well-being language that accompanies them. Whether campaign periods alter positive affect and hope, and whether happiness aligns with action language, remains unresolved. We analysed 364,118 public Twitter/X posts from Earth Day, Earth Hour, Global Climate Action Day and World Environment Day in 19 occurrence-years, using 30-day pre-event, event and post-event windows. A versioned weighted lexical model estimated happiness, future-oriented hope, collective capability, distress and action language. Event-period happiness prevalence was 9.02 percentage points higher than the pre-event baseline (95\% CI 1.32--16.73; nominal $p=0.0218$; BH $q=0.1516$), whereas paired occurrence contrasts showed a 10.75-point decline in action language, indicating a happiness--action divergence. The happiness estimate remained positive across composition and text-deduplication checks, but was less precise under a 19-cluster wild bootstrap ($p=0.0565$). Happier source posts had lower odds of an observed matched retweet cascade (OR 0.457, 95\% CI 0.233--0.896);
}

\keywords{climate communication, expressed well-being, happiness, hope, collective efficacy, Twitter, retweet cascades, computational social science}

\maketitle

\section{Introduction}\label{sec:introduction}

Climate change is a biophysical threat, a public-health problem and a collective communication problem. Recent assessments describe effects on physical health, displacement, livelihoods and mental health, while emphasizing that vulnerability is socially patterned and that responses depend on institutions and collective capacity\cite{romanello2023countdown,lawrance2022mental,ostrom2009framework}. Public communication is part of this response system. It transmits information about temperature, emissions and policy, but it also signals whether a risk is shared, whether a desirable future remains imaginable and whether people can act together. These cues matter because climate concern is widespread while sustained attention and action are uneven.

Emotion research treats emotions as appraisal processes that organize attention, meaning and action readiness. Several emotions can occur in the same episode or utterance\cite{scherer2005emotions,lazarus1991adaptation,gross1998regulation}. Positive emotions can broaden thought and support social resources, while positive affect can coexist with danger, dissatisfaction or inaction\cite{fredrickson2001positive}. Climate communication research has moved beyond a simple fear-versus-hope opposition. Concern, grief, anger, moral outrage, solidarity, hope and efficacy can coexist, and people with different identities or experiences can read the same message differently\cite{chapman2017reassessing,markowitz2014psychology,brosch2021affect}.

This distinction matters for work on well-being. Climate anxiety is associated with persistent worry, impairment and uncertainty about institutional responses; climate-related worry also encompasses experiences beyond a mental-health disorder\cite{clayton2020anxiety,pihkala2020anxiety,hickman2021anxiety}. Population studies also show that climate concern can coexist with hope, meaning and engagement. Among young people, coping strategies that combine hope with problem-focused engagement are related to environmental action and well-being, whereas guilt, denial or helplessness can undermine both\cite{ojala2012coping,ojala2012hope,ojala2015hope}. Hope is multidimensional. Snyder's hope theory separates agency, the will to pursue a goal, from pathways, the routes perceived to reach it\cite{snyder2002hope}. In climate contexts, constructive hope and passive optimism can lead to different forms of language and action\cite{cohenchen2019hope,marlon2019mobilization,mortreux2025hope}.

Evidence about climate-message framing remains mixed. Images and text that communicate attainable solutions can increase hope and policy support, but progress-focused messages can also weaken motivation when they make the problem appear solved or when they fail to name responsibility and agency\cite{feldman2018hope,hornsey2016hope,morris2020endings}. Recent preregistered and large-scale message studies likewise report heterogeneous effects across emotional frames\cite{lammers2026communicating,voelkel2026megastudy}. The implication is methodological as well as theoretical: a study should measure positive affect, future-oriented hope and action language separately, and should test whether their temporal patterns converge or diverge. A single sentiment score can obscure precisely the distinction that climate-communication theory seeks to explain.

Collective efficacy provides a bridge between emotion and action. Social identity models of collective action predict that people act when they perceive a shared grievance, identify with a group and believe that coordinated effort can produce change\cite{vanzomeren2008integrative,vanZomeren2012perspective}. The social identity model of pro-environmental action further proposes that group norms, efficacy beliefs, environmental appraisal and perceived costs jointly shape action\cite{fritsche2018simpea}. Empirical studies link collective efficacy to pro-environmental intentions through both self-efficacy and group identification\cite{bamberg2015collective,jugert2016efficacy,reese2019social}. The theory of planned behaviour and value-belief-norm approaches add behavioural intentions, perceived control and moral obligation to this account\cite{ajzen1991planned,stern2000environmental,fielding2008planned}. In text, collective efficacy concerns shared ability, coordination, institutional leverage and a community's capacity to produce a climate outcome.

Well-being and action may follow different trajectories. Positive language can support participation, mark solidarity, express a rhetorical position or accompany low-cost symbolic involvement. Anger or distress can motivate action when paired with efficacy and a target, but can also produce withdrawal when paired with futility\cite{gifford2011dragons,cohenchen2019hope,fritsche2018simpea}. Surveys and experiments have measured these mechanisms in controlled settings. They are needed for psychological validity, but they usually expose participants to researcher-selected messages and sample smaller populations.

Social media provide a second source of evidence. Twitter/X posts are public, time-stamped and connected to retweet and reply metadata, so researchers can study the timing and visibility of campaign discourse. Computational social science has shown that digital traces can reveal aggregate temporal patterns that are difficult to observe with conventional surveys\cite{lazer2009computational,gentzkow2019text,grimmer2013promise}. Hedonometer-style dictionaries estimate word pleasantness and have been used to track expressed happiness in songs, blogs and Twitter streams\cite{dodds2010happiness,dodds2011twitter}. LIWC supplies psychologically motivated categories such as positive emotion, negative emotion, social processes and agency\cite{tausczik2010liwc,pennebaker2007liwc}. These tools produce reproducible language measures. 

Social-media aggregates have several limitations. Word meanings depend on context, and automated lexicons can reproduce cultural and demographic biases\cite{caliskan2017semantics,hutto2014vader}. User activity is unequal, platform audiences are selected, and a burst of posts may be driven by a small group of highly active accounts. Seasonal rhythms and news cycles can create apparent mood changes unrelated to a campaign\cite{golder2011diurnal}. Keyword rules can overrepresent organized communities and miss indirect or image-based communication. Big-data studies also fail when a digital proxy is treated as the construct itself, as the Google Flu case illustrates\cite{lazer2014paradigm}. We address these limits by freezing the corpus before outcome analysis, fixing the campaign rules, measuring several constructs, weighting and deduplicating authors in sensitivity analyses and reporting uncertainty with 19 occurrence clusters.

The network analysis has a separate limitation. Retweets create observable source--child links, but a retweet usually reproduces the source text, so source and child receive the same lexical score. A retweet edge records exposure or cascade selection; recipient emotional uptake remains unobserved. Experiments have shown that social-network exposure can alter emotional expression under particular platform conditions\cite{kramer2014contagion}, while observational studies link emotion and moral language to diffusion\cite{brady2017emotion}. Exposure is also shaped by algorithmic ranking, homophily and ideological selection\cite{bakshy2015exposure,bail2018exposure,vosoughi2018false}. We therefore model whether a matched cascade is observed and, conditional on observation, its size and timing.

We focus on recurring climate campaigns. Earth Day, Earth Hour, Global Climate Action Day and World Environment Day provide repeated temporal anchors with different institutional histories, participation styles and levels of public visibility. The repeated-event design allows within-occurrence contrasts while preserving differences across campaign families and years.

The temporal design also sets the limits of the comparison. A pre-event window provides a local reference for the same occurrence, the event window covers the period when campaign cues are most salient, and the post-event window tests persistence after the focal date. The design remains vulnerable to coincident news, seasonal routines and changes in account composition. We treat the pre/event/post contrast as an observational comparison. A sign that recurs across occurrence-years suggests a repeated discourse pattern; a sign found in one campaign is event-specific.

The design distinguishes three levels of inference. At the post level, a score describes one message. At the occurrence-day level, a prevalence describes the composition of messages on a day. At the occurrence-year level, a paired contrast describes change relative to that occurrence's local baseline. These measures describe messages and their temporal composition, leaving individual psychological states unobserved. A campaign can change who posts, which organizations are visible or which genres are shared while readers' well-being remains unchanged. We report the unit of inference with each estimate to avoid reading an ecological association as an individual effect.

These considerations motivate three connected research questions. \textbf{RQ1.} Do recurring climate campaigns coincide with changes in expressed happiness, hope, distress and collective-capability language across the pre-event, event and post-event periods? \textbf{RQ2.} If happiness language changes, does it move in parallel with explicit climate-action language, or do campaign periods show a measurable happiness--action decoupling? \textbf{RQ3.} Finally, how is happiness-labelled climate discourse represented in observed retweet cascades across the same periods?

We estimate changes in expressed well-being language in a versioned corpus and measure how those signals move with action language. External lexical controls and a reproducibility audit assess measurement stability. Paired occurrence contrasts, small-cluster resampling and linkage checks assess inferential sensitivity. The main theoretical possibility is a happiness--action difference: campaign discourse may contain more positive or future-oriented language without a parallel rise in explicit action language.


\section{Methods}\label{sec:methods}

\subsection{Data Collection}\label{subsec:data}

The analysis uses a versioned local archive of public Twitter posts collected by the project team. The source archive contains approximately 170 million posts retrieved from November 1, 2010 to November 30, 2022 with the broad climate-query terms ``climate change'', ``global warming'', ``climate crisis'', ``climate issue'', ``carbon neutrality'' and ``low-carbon''. From this source archive, the formal analytic table contains 364,118 posts from four pre-specified climate campaign families: Earth Day, Earth Hour, Global Climate Action Day and World Environment Day. 
Campaign membership is assigned by an outcome-blind keyword/hashtag scanner defined before the well-being outcomes were inspected. The scanner relies exclusively on exact and normalized campaign names and their documented hashtags. Posts are retained with a stable post identifier, author identifier, timestamp, text, language, campaign family, occurrence-year and period label. Retweet, reply and quote identifiers are retained for the separate network audit. Supporting Information, Table~S1, provides the frozen data design and analysis units.

Each occurrence has a 30-day pre-event window, an event window defined by the campaign's operational date range, and a 30-day post-event window. The event boundaries are fixed in the frozen manifest and are not optimized to maximize a result. The primary temporal unit is an occurrence-day, producing 1,008 daily observations; paired change analyses use the 19 occurrence-years as the independent occurrence units. This repeated-event design allows the same campaign family to be observed across years while preserving differences in scale and institutional context. Occurrence-year is the unit of inference.

The large input files were scanned in a columnar, chunked workflow. Only the fields needed for campaign assignment, text scoring, temporal aggregation and linkage were read into memory. Exact duplicate identifiers were removed before aggregation, and text-template deduplication is used only as a pre-specified robustness check. The frozen manifest records the input file hashes, row counts, column selections, campaign rules and scoring version, so all main tables can be regenerated without loading the complete archive into memory.

\subsection{Constructs and text-level measurement}\label{subsec:measurement}

The primary construct is \emph{expressed happiness}: whether a post contains a weighted lexical signal associated with pleasant affect, positive appraisal or social enjoyment. Secondary constructs are hope-waypower, collective capability, distress and action language. Hope-waypower represents future-oriented improvement coupled with a route, effort or feasible change. Collective capability represents language about shared ability, coordination, institutions or a community's capacity to produce a climate outcome. Distress captures threat, worry, loss and helplessness language. Action captures explicit behavioural, political or policy verbs. These are text-level language measures; they provide no diagnosis, trait measure or direct reading of subjective well-being.

For post $i$ and construct $k$, the frozen scorer computes a weighted lexical score $s_{ik}=\sum_{w\in i}n_{iw}v_{wk}$, where $n_{iw}$ is the token count and $v_{wk}$ is the versioned token weight. The primary outcome is a binary signal $I_{ik}=1(s_{ik}>0)$, and the continuous score is retained for sensitivity analyses. Token contributions and dictionary versions are stored with each score, allowing a post-level result to be audited without re-estimating the weights. Scores are aggregated within occurrence-day as prevalence, $Y_{dk}=N_d^{-1}\sum_i I_{ik}$, where $N_d$ is the number of retained posts on day $d$. Supporting Information, Table~S2, gives the operational definitions and interpretation boundaries for all five constructs.

The lexical model was constructed from psychologically motivated seed dictionaries and corpus-derived weights, with action terms kept separate from the outcome construct. The happiness score excludes action terms, and the action score excludes happiness terms. LIWC-2007 positive-emotion prevalence and labMT pleasantness are external lexical controls. Their correlations with the primary score provide convergent language evidence; shared text and vocabulary limitations preclude independent criterion validation.

To assess scoring reproducibility, two independent model coders apply the frozen codebook to a random set of 451 English posts while ignoring the existing scores. Agreement is summarized with nominal Krippendorff's alpha and Cohen's kappa. This audit measures the consistency of codebook application and offers no human validation of the text scores as measures of a user's psychological state.

\subsection{Primary temporal model and estimands}\label{subsec:model}

The primary estimand is the event-period change in happiness prevalence relative to the pre-event period. We fit a linear-probability occurrence-day model:

\begin{equation}
\begin{aligned}
H_d={}&\beta_0+\beta_1(\mathrm{Event}_d\times\mathrm{Campaign}_d)+\beta_2\mathrm{Post}_d\\
&+\gamma\mathrm{DayOfWeek}_d+\delta\log(1+N_d)+\eta\mathrm{Year}_d+\varepsilon_d,
\end{aligned}
\label{eq:primary}
\end{equation}

where $H_d$ is daily happiness prevalence, $N_d$ is the daily post count, and campaign-family indicators allow the period contrast to vary across families. Standard errors are clustered by occurrence-year. We report percentage-point estimates, 95\% confidence intervals, nominal $p$ values and Benjamini--Hochberg adjusted values for the pre-specified model family\cite{benjamini1995false}. A logistic model for post-level presence and a beta-binomial sensitivity for daily counts are included in Supporting Information.

The event-study specification adds relative day, event-step and pre-event trend terms. It characterizes temporal shape without invoking a sharp causal discontinuity. The occurrence-level paired analysis computes pre-to-event and pre-to-post changes for each of the 19 occurrence-years. Because the number of clusters is small, we use conventional cluster-robust standard errors, a Rademacher wild-cluster bootstrap and exact sign-flip tests. The sign-flip tests permute the sign of each occurrence contrast and preserve the repeated-event structure. The sensitivity family excludes retweets, restricts to English non-retweets, weights authors equally within day, trims authors above the frozen 99th-percentile activity cutoff, and removes exact or normalized duplicate text.

\subsection{Direct happiness--action decoupling}\label{subsec:index}

For occurrence $o$ and contrast $c\in\{\mathrm{event\!\! -\!\! pre},\mathrm{post\!\! -\!\! pre}\}$, the direct decoupling index is

\begin{equation}
D_{o,c}=\left(H_{o,c}-H_{o,\mathrm{pre}}\right)-\left(A_{o,c}-A_{o,\mathrm{pre}}\right),
\label{eq:decoupling}
\end{equation}

where $A$ is action-language prevalence. A positive value means that happiness language increased relative to action language. We report the distribution of $D$ across occurrence-years, the number of positive contrasts, exact sign-flip $p$ values and Wilcoxon signed-rank sensitivities with BH adjustment. Supporting Information, Table~S3, reports the occurrence-level paired contrasts. This descriptive divergence estimand addresses neither mediation nor causation, and it leaves symbolic participation and changes in campaign-talk composition indistinguishable.

\subsection{Observed retweet cascades}\label{subsec:retweet}

The network analysis uses retained retweet and quote identifiers to match child posts to in-corpus source posts. The linkage audit first reports unresolved identifiers and identifiers affected by scientific-notation precision loss. Only edges with a resolvable source and a common campaign occurrence are used for the formal analysis. The primary network estimand is the association between source-post happiness and the observation of a matched cascade. A hurdle model separates (i) the probability that any matched retweet is observed and (ii) the conditional count of matched retweets, with occurrence-clustered uncertainty. A source-level model of log time to first matched retweet serves as a timing sensitivity, with right censoring left uncorrected.

The data-generating process sets the interpretation: a retweet copies the source text, so source and child happiness scores are identical on matched edges. The network model traces exposure and cascade selection. An emotional-contagion claim would require an independent recipient outcome, an exposure denominator, reliable source--child linkage and a design that addresses homophily, ranking and confounding\cite{kramer2014contagion,brady2017emotion,bakshy2015exposure,vosoughi2018false}.

\subsection{Missingness, multiplicity and quality controls}\label{subsec:quality}

Missing text, malformed timestamps and rows without a stable identifier are removed before scoring and are counted in the frozen data-quality table. Missing text and construct scores remain unimputed. Daily prevalence is calculated from the posts observed in that day; therefore, a low-volume day contributes a noisier estimate and adjacent dates supply no replacement values. Volume enters the primary model as a covariate, and author-equal and text-deduplicated variants test whether high-volume accounts or repeated templates drive the result. Campaign rules, event boundaries and the 99th-percentile activity cutoff are stored before the final outcome table is generated.

The confirmatory family contains the event-period happiness estimate, the pre-specified event-study step and the two direct decoupling contrasts. Secondary constructs, campaign-specific estimates, network hurdles and M3 temporal-precedence scans are labelled exploratory or Supporting Information analyses. Benjamini--Hochberg correction is applied within each declared family. We report effect sizes and intervals even when adjusted or small-cluster $p$ values fail conventional significance thresholds. This prevents lexical and account-group comparisons from being treated as independent discoveries.

\subsection{Exploratory account-group analysis and reproducibility}\label{subsec:m3}

M3 text-only predictions\cite{wang2019demographic} are retained for an exploratory Supporting Information analysis of aggregate account groups. Age and gender labels come from the model; the archive contains no self-reports or verified demographics. Sparse categories are collapsed or excluded under pre-specified rules. Granger-style tests use first-differenced daily happiness series and a three-day lag to test temporal precedence, with BH correction within the testing family\cite{granger1969investigating}. Supporting Information, Fig.~S1, reports the complete account-group temporal-precedence scan. These tests provide no causal evidence and sit outside the primary happiness claim.

All reported numbers are linked to machine-readable CSV/JSON tables, a frozen manifest and a dependency-free audit. Refresh scripts resolve the analysis root from the file location or the environment variable \texttt{ANALYSIS\_ROOT}; the formal input is supplied through \texttt{FORMAL\_PARQUET}. Raw post text and user-level exports fall outside the redistributable release. The code release contains the scoring, aggregation, model and figure scripts, while reproduction of the full text-level analysis requires an authorized local archive. The reproducibility code, derived intermediate data and publication figures are archived at Figshare: \url{https://doi.org/10.6084/m9.figshare.33207624}.

\section{Results}\label{sec:results}

\subsection{RQ1: campaign-period changes in expressed well-being}\label{subsec:temporal}

\begin{figure*}[t]
\centering
\includegraphics[width=0.92\textwidth]{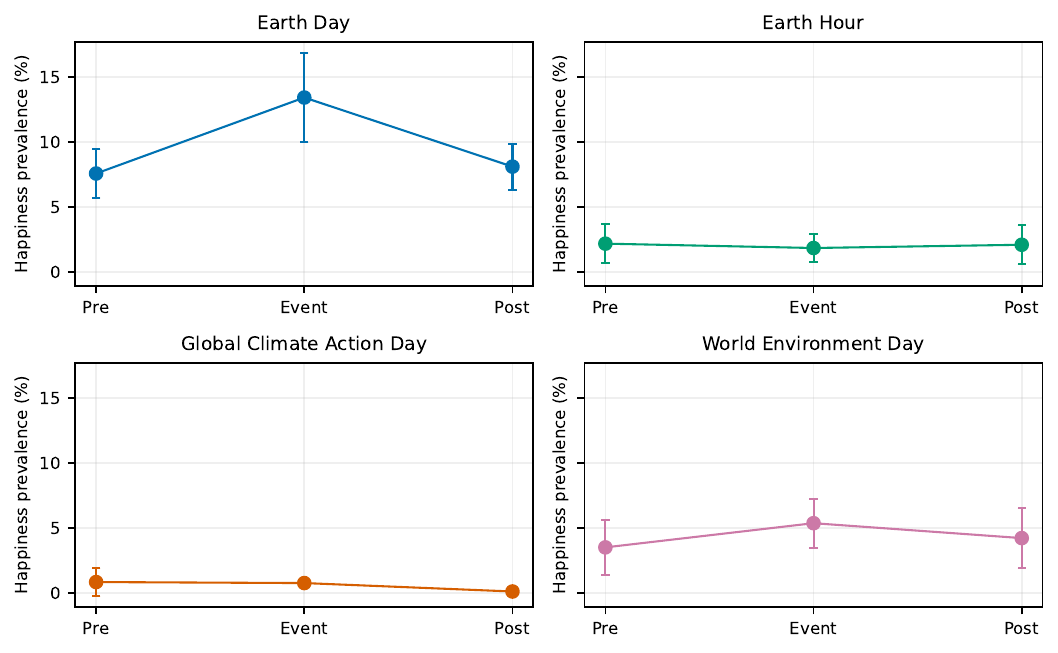}
\caption{Expressed-happiness prevalence across campaign families and temporal periods. Points and intervals summarize the frozen occurrence-period cells; the figure reports aggregate language prevalence.}\label{fig:campaign}
\end{figure*}

Figure~\ref{fig:campaign} shows the campaign-period pattern across the four families. In the primary linear-probability model, the event-period coefficient is $+9.02$ percentage points (95\% CI $1.32$--$16.73$; nominal $p=0.0218$). The corresponding BH value for the model family is $q=0.1516$, and we report the finding as an uncertain positive adjusted association. The post-period coefficient is $+0.69$ percentage points (95\% CI $-6.82$--$8.19$; $p=0.858$). The segmented event-study finds no pooled linear pre-event trend ($-0.019$ percentage points per relative day; $p=0.848$) and an event step of $+5.41$ percentage points ($p=0.056$). Supporting Information, Table~S4, provides the complete temporal estimates and small-cluster checks.

\begin{figure*}[t]
\centering
\includegraphics[width=1\textwidth]{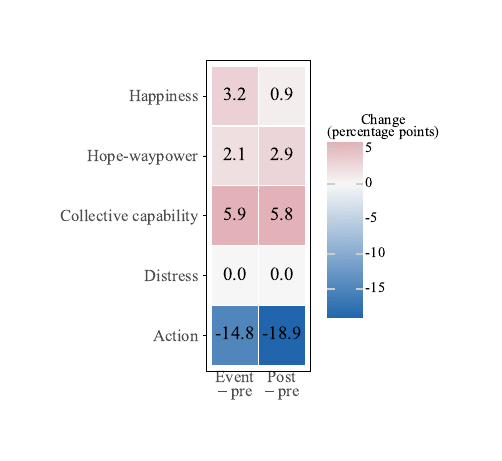}
\caption{Construct changes across occurrence-years. The heatmap displays paired period contrasts for happiness, hope-waypower, collective capability, distress and action language. The secondary constructs are lexical indicators and are interpreted more cautiously than happiness.}\label{fig:heatmap}
\end{figure*}

At the paired occurrence-year level, happiness prevalence increases by 4.94 percentage points from pre-event to event and by 3.39 percentage points from pre-event to post-event. The construct-change summary in Figure~\ref{fig:heatmap} shows that hope-waypower and collective-capability language also move upward descriptively, while distress remains close to zero in the aggregate contrasts. Campaign-specific estimates are heterogeneous, precluding any single campaign from serving as a universal proxy for climate communication.

\subsection{RQ2: happiness--action decoupling}\label{subsec:decoupling}

For each occurrence-year, we calculated the change in happiness prevalence and the change in action prevalence relative to the pre-event period. The direct decoupling index is $D=\Delta H-\Delta A$. During the event contrast, mean happiness increases by 4.94 percentage points while action language decreases by 10.75 percentage points, giving a mean divergence of 15.69 percentage points. The divergence is positive in 16 of 19 occurrence-years. For the post contrast, the corresponding changes are +3.39 and $-14.03$ percentage points, with a mean divergence of 17.42 percentage points and positive divergence in 17 of 19 occurrence-years. Exact occurrence-level sign-flip tests are below 0.011 for the six pre-specified happiness, action and decoupling contrasts; Supporting Information, Table~S3, reports these contrasts in full.

Within the frozen campaign windows, happiness-labelled language and explicit action language show divergent trajectories. The pattern is compatible with several mechanisms, including symbolic participation, collective coping and changes in the composition of campaign talk. The design leaves the contribution of these mechanisms unidentified, and causal claims about positive language and action lie beyond its scope.

\subsection{Measurement checks and robustness}\label{subsec:robustness}

Happiness mean correlates with the labMT happiness mean ($r=0.412$, bootstrap 95\% CI $0.378$--$0.442$) and with LIWC positive-emotion proportion ($r=0.551$, bootstrap 95\% CI $0.491$--$0.597$). These convergent lexical checks share a language basis, which limits their use for criterion validation. Supporting Information, Table~S5 and Fig.~S2, provide the full external lexical comparisons.

\begin{figure*}[t]
\centering
\includegraphics[width=0.92\textwidth]{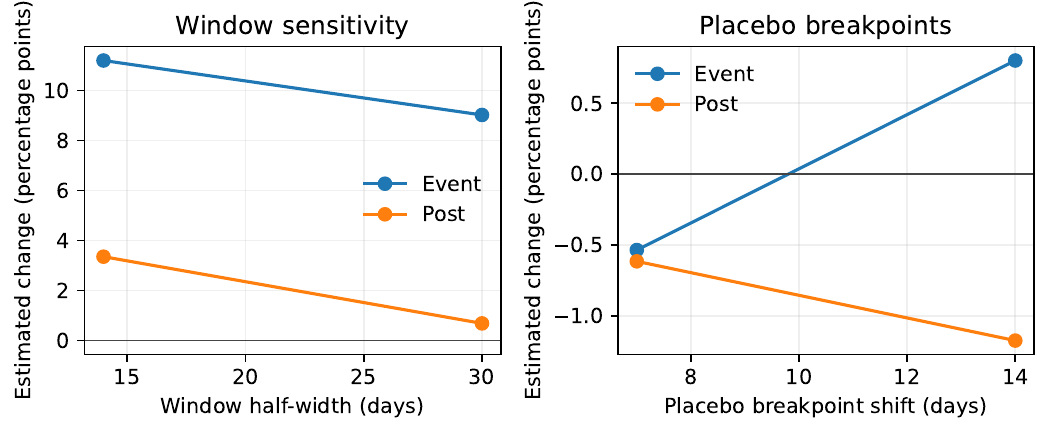}
\caption{Robustness and placebo checks for the event-period happiness association. The direction is stable across composition and text-template variants, while the small-cluster wild-bootstrap result is appropriately more conservative.}\label{fig:robustness}
\end{figure*}

The event-period estimate remains positive across every pre-specified composition and text-template sensitivity, ranging from +8.43 to +9.04 percentage points. Figure~\ref{fig:robustness} summarizes these checks, and Supporting Information, Table~S6, gives the full numerical results. A Rademacher wild-cluster bootstrap over 19 occurrence-years gives $p=0.0565$ for the common event-period coefficient. The wild-bootstrap result is less precise than the conventional clustered result.

Two independent model coders annotated the same 451 English posts using the frozen instructions. Nominal Krippendorff's alpha is 0.811 for happiness, 0.519 for distress, 0.321 for hope, 0.374 for collective capability and 0.344 for action. Supporting Information, Table~S7, gives the full annotation audit. The audit documents reproducibility of the coding instructions, especially for happiness; human criterion validity remains unestablished. The lower agreement for the secondary constructs motivates their treatment as lexical indicators with limited psychological interpretation.

\subsection{RQ3: observed retweet-cascade exposure}\label{subsec:network}

Many source identifiers in the formal edge table remain unresolved, including identifiers affected by scientific-notation precision loss; matched cascade counts are therefore lower bounds. Supporting Information, Table~S8, provides the complete linkage-coverage audit. A hurdle model separates whether any matched cascade is observed from the number of matched retweets conditional on a positive count. Happier source posts have lower odds of any observed matched cascade (OR 0.457, 95\% CI 0.233--0.896; clustered $p=0.0225$). Conditional cascade size is directionally similar but uncertain after occurrence clustering (IRR 0.302, 95\% CI 0.047--1.928; $p=0.206$). The timing sensitivity yields $\beta=0.089$ for log time to the first matched retweet (95\% CI $-0.190$--$0.368$; $p=0.532$).

\begin{figure*}[t]
\centering
\includegraphics[width=0.92\textwidth]{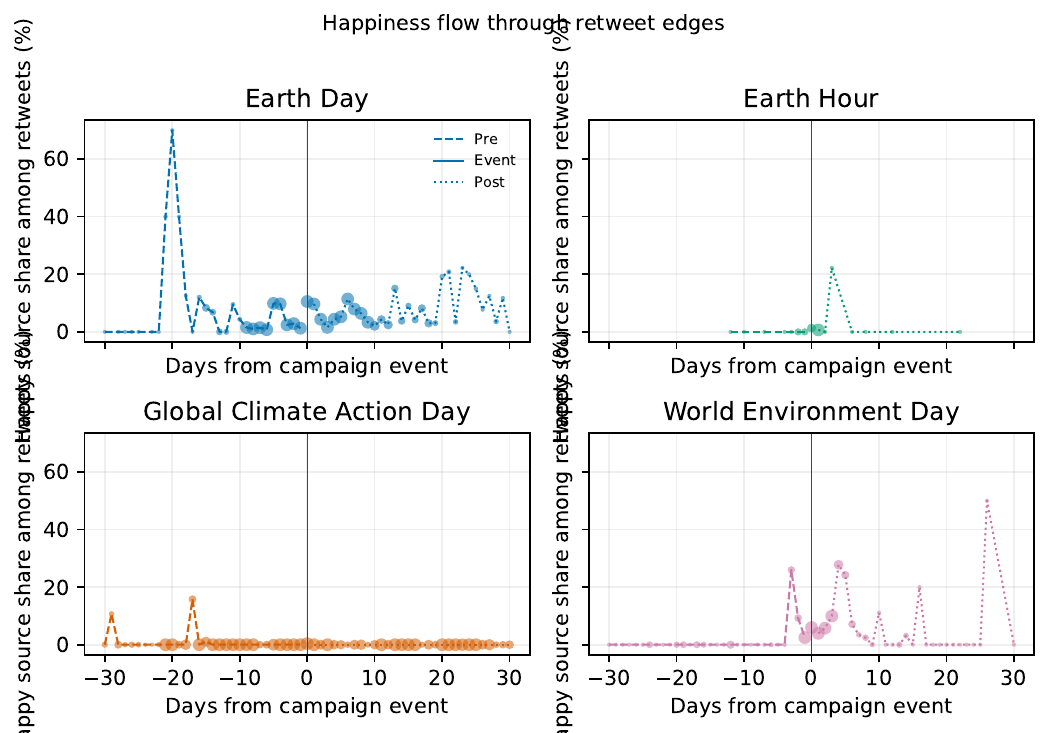}
\caption{Observed retweet-edge representation of happiness-labelled climate discourse. The figure reports matched exposure and cascade structure; duplicated retweet text leaves changes in recipient emotion unobserved.}\label{fig:flow}
\end{figure*}

The same text is copied by a retweet, and source and child happiness scores are identical for every matched edge. The matched data contain no independent recipient emotional outcome. Figure~\ref{fig:flow} therefore reports observed exposure and cascade selection, with emotional contagion and causal diffusion outside the estimand.

\FloatBarrier
\clearpage

\section{Discussion}\label{sec:discussion}

The main finding is a difference between two forms of climate discourse. Happiness-labelled language rises around campaign periods in the primary panel, while explicit action language decreases in paired occurrence contrasts. A campaign can coincide with more positive-affect or social-possibility language without a corresponding increase in explicit action language. These data describe an observed discourse pattern; causal effects of campaigns on happiness or action remain outside the analysis.

The result is consistent with theories that distinguish constructive hope from passive optimism. Hope can signal that a desirable future is imaginable, but its relationship with action depends on whether the text also names agency, pathways and collective capacity\cite{ojala2012hope,marlon2019mobilization}. Our lexical indicators cannot recover the full psychological appraisal or lived context behind a post. They can identify when these semantic components co-occur or diverge in a large public corpus. A happiness word records wording in a post; the user's happiness remains unobserved.

The network analysis answers a narrower question. Happiness-labelled source posts are less likely to have an observed matched cascade in the hurdle model, but the conditional cascade-size and timing estimates are uncertain. Source and child texts are identical, and the graph contains no independent recipient outcome after exposure. The analysis therefore concerns selection into observed cascades and the visibility of source language. Causal diffusion claims would require independent recipient outcomes, an exposure denominator, reliable source identifiers and a design that addresses confounding and censoring.

Five limitations qualify these results. First, the design is observational and contains no randomized campaign exposure. The event-period coefficient is an adjusted temporal association, and the wild-cluster sensitivity is borderline with only 19 occurrence-years. Second, campaign families were selected through observed keywords and hashtags, which confines the corpus to campaign-labelled Twitter/X discourse. Third, the happiness, hope, collective-capability, distress and action measures are lexical/model-inferred indicators. LIWC and labMT provide convergent language checks, and the two-model audit tests reproducibility, but neither establishes human criterion validity. Fourth, the retweet network is incomplete because source identifiers are unresolved for many rows and the outcome counts only matched edges. Fifth, M3 age and gender outputs are model-inferred aggregate groups with possible systematic error; the associated temporal-precedence scans remain exploratory Supporting Information results.

The analysis provides a reproducible way to study positive well-being language in climate communication while keeping social-media text separate from clinical measurement. Its contribution is methodological and descriptive: repeated campaign windows, paired happiness--action contrasts, lexical controls, author and text-template sensitivities, small-cluster inference and a network-linkage audit. Future work should add independently validated human annotations, broader non-campaign comparison periods, multilingual measurement and longitudinal designs that distinguish affective expression from individual well-being and behavioral change.

\backmatter

\bmhead{Data and code availability}

Reproducibility code, the dependency specification, final statistical audit and machine-readable derived outputs are archived at Figshare: \url{https://doi.org/10.6084/m9.figshare.33207624}. Raw Twitter/X text and the full user-level archive are excluded from redistribution. 
\bibliography{references}

\end{document}


\title[Supporting Information]{Supporting Information for: AI-inferred expressed well-being and collective-action discourse}
\author*{\fnm{Wentao} \sur{Xu} \textnormal{(Doctor of Informatics)}}\email{myrainbowandsky@gmail.com}

\maketitle


\section{Scope and interpretation}\label{sec:scope}

The Supporting Information (SI) documents the data freeze, construct scoring, model specifications, sensitivity checks, external lexical comparisons, retweet-linkage audit and exploratory account-group temporal-precedence analyses. It records the analyses referenced in the main manuscript, with confirmatory hypotheses fixed before the main results. The primary estimand remains the change in the prevalence of expressed happiness language during campaign periods. All scores are properties of public text; they provide no clinical measure, survey response or direct observation of an author's subjective well-being.

The SI separates confirmatory and exploratory material. The confirmatory family contains the event-period happiness association, the segmented event-study step and the two occurrence-level happiness--action contrasts. Campaign-specific estimates, secondary constructs, model-annotation agreement, retweet hurdles and M3 age/gender scans are reported as secondary or exploratory. Benjamini--Hochberg correction is applied within the declared testing families\cite{benjamini1995false}. With only 19 occurrence-years, wild-cluster and exact sign-flip results are reported alongside conventional asymptotic $p$ values.

\section{Data freeze and scoring protocol}\label{sec:data}

\subsection{Frozen corpus and event windows}\label{subsec:freeze}

The formal frozen table contains 364,118 public Twitter/X posts from 19 occurrence-years and four campaign families: Earth Day, Earth Hour, Global Climate Action Day and World Environment Day. Each occurrence has a 30-day pre-event window, the operational event window, and a 30-day post-event window. Campaign membership was assigned by the pre-specified keyword/hashtag scanner before outcome analysis. The primary unit is the occurrence-day (1,008 daily observations); paired changes use the 19 occurrence-years as the independent units.

\begin{table}[t]
\caption{Frozen data design and analysis units.}\label{tab:S1-design}
\centering\small
\begin{tabularx}{\linewidth}{>{\raggedright\arraybackslash}p{0.31\linewidth}X}
\toprule
Item & Definition \\
\midrule
Formal corpus & 364,118 public posts; 364,036 globally unique post identifiers \\
Campaign families & Earth Day, Earth Hour, Global Climate Action Day and World Environment Day \\
Repeated events & 19 campaign occurrence-years \\
Temporal windows & 30-day pre-event, operational event window and 30-day post-event \\
Primary panel & 1,008 occurrence-day observations \\
Paired unit & Occurrence-year; $n=19$ \\
Primary outcome & Daily prevalence of a happiness-language signal \\
Secondary outcomes & Hope-waypower, collective capability, distress and action language \\
\bottomrule
\end{tabularx}
\end{table}

\subsection{Construct definitions and audit trail}\label{subsec:constructs}

Scoring follows the frozen procedure described in the main manuscript. Token contributions and dictionary versions are retained with each score, allowing the text-level audit to trace an individual result to its contributing words.

\begin{table}[t]
\caption{Operational definitions used in the text-level scorer.}\label{tab:S2-constructs}
\centering\small
\begin{tabularx}{\linewidth}{>{\raggedright\arraybackslash}p{0.24\linewidth}X>{\raggedright\arraybackslash}p{0.23\linewidth}}
\toprule
Construct & Operational meaning in text & Interpretation boundary \\
\midrule
Happiness & Pleasant affect, positive appraisal or social enjoyment & Expressed happiness language, not subjective happiness \\
Hope-waypower & Desired improvement coupled with a route, effort or feasible change & Future-oriented discourse, not a hope-trait score \\
Collective capability & Shared ability, coordination, institutions or group capacity & Expressed collective efficacy language, not verified efficacy beliefs \\
Distress & Threat, worry, loss, grief or helplessness language & Expressed distress language, not a mental-health diagnosis \\
Action & Explicit behavioural, political or policy verbs & Action-oriented discourse, not observed behaviour \\
\bottomrule
\end{tabularx}
\end{table}

LIWC-2007 positive-emotion categories and labMT pleasantness are retained as external lexical comparisons\cite{tausczik2010liwc,dodds2010happiness,dodds2011twitter}. Because all instruments operate on the same text, their correlations provide convergent language checks; independent criterion validity remains unavailable.

\section{Paired happiness--action contrasts}\label{sec:si-decoupling}

Table~S3 reports the occurrence-level happiness--action contrasts calculated with the decoupling index defined in the main manuscript.

\begin{table}[t]
\caption{Direct happiness--action decoupling by occurrence-year.}\label{tab:S3-decoupling}
\centering\small
\begin{tabular}{lrrrrrr}
\toprule
Contrast & $n$ & $\Delta H$ (pp) & $\Delta A$ (pp) & $D$ (pp) & Positive/$n$ & Wilcoxon $q$ \\
\midrule
Event--pre & 19 & 4.944 & -10.75 & 15.69 & 16/19 & 0.000267 \\
Post--pre & 19 & 3.392 & -14.03 & 17.42 & 17/19 & 0.000267 \\
\bottomrule
\end{tabular}
\end{table}

Exact occurrence-level sign-flip tests for the six pre-specified happiness, action and decoupling contrasts are below 0.011. These tests establish a recurrent divergence in the frozen discourse panel; the underlying psychological mechanisms and causal effects remain unidentified.

\section{Exploratory account-group temporal precedence}\label{sec:m3}

M3 text-only predictions provide exploratory age and gender account groups. The analysis uses first-differenced daily happiness series and a three-day lag, with eligibility rules and BH correction defined before inspecting the direction of the results. Across 504 tested direction--lag records, 36 were primary eligible tests and none was significant after BH correction ($\min q_{\mathrm{primary}}=0.165$). Eight tests were flagged at some non-primary lag. They remain exploratory and sit outside the main causal narrative.

\begin{figure*}[t]
\centering
\includegraphics[width=0.94\textwidth]{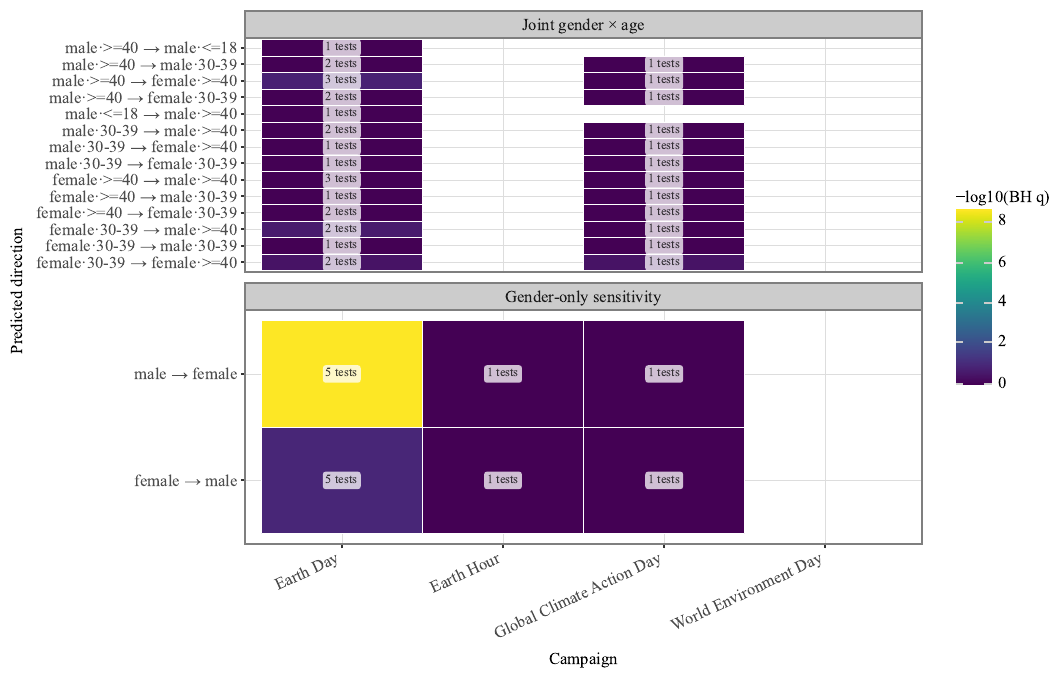}
\caption{M3-inferred account-group temporal precedence in happiness scores. Tiles show the smallest BH-adjusted $q$ value among primary three-day first-difference tests for each campaign and direction. All eligible tests, including null results, are retained. A temporal prediction relation provides no causal effect estimate, and the account groups are model-derived.}\label{fig:S1-granger}
\end{figure*}

The age/gender results serve as a stress test of predictable temporal ordering in aggregate account-group series. Causal relations among demographic groups lie outside this analysis. The complete CSV retains every campaign, direction, lag, transformation, sample size, $F$ statistic and adjusted value.

\section{Temporal models and primary results}\label{sec:temporal}

\subsection{Model specification}\label{subsec:temporal-model}

The primary occurrence-day linear-probability model is specified in the main manuscript. Here, we report the complete temporal estimates and small-cluster checks. Standard errors are clustered by occurrence-year, and the event coefficient is reported in percentage points. A segmented event-study adds relative day, an event step and period-specific slopes. No causal discontinuity is assumed: the design is an observational pre/event/post comparison vulnerable to coincident news, seasonality and changing account composition.

\begin{table}[t]
\caption{Primary temporal estimates and small-cluster checks.}\label{tab:S4-temporal}
\centering\small
\begin{tabularx}{\linewidth}{>{\raggedright\arraybackslash}p{0.31\linewidth}p{0.16\linewidth}p{0.20\linewidth}X}
\toprule
Estimand & Estimate (pp) & 95\% CI & $p$ / inference \\
\midrule
Event period, primary model & 9.02 & 1.32--16.73 & 0.0218; 19-cluster CRSE \\
Post period, primary model & 0.69 & -6.82--8.19 & 0.858; 19-cluster CRSE \\
Pooled pre-event slope & -0.019/day & -0.209--0.172 & 0.848; event study \\
Event step, segmented study & 5.41 & -0.14--10.96 & 0.0561; event study \\
Wild-cluster event coefficient & 9.02 & --- & 0.0565; Rademacher $p$ \\
\bottomrule
\end{tabularx}
\end{table}

The conventional event-period association is positive, but its BH-adjusted value is $q=0.1516$ and the wild-cluster result is borderline. We report a positive, uncertain temporal association; campaign causation remains unestablished.

\section{Measurement validation and external controls}\label{sec:validation}

\subsection{LIWC and labMT comparisons}\label{subsec:external}

The primary happiness prevalence correlates with labMT positive-word share (Pearson $r=0.354$, $p=0.0069$; Spearman $\rho=0.373$, $p=0.0043$) and with LIWC positive-emotion prevalence (Pearson $r=0.565$, $p<0.00001$; Spearman $\rho=0.578$, $p<0.00001$). The labMT happiness-mean comparison is weaker (Pearson $r=0.338$, $p=0.0102$; Spearman $\rho=0.200$, $p=0.136$), so dictionary agreement serves as a sensitivity check. Collective capability shows positive convergent evidence with LIWC social-process language (Spearman $\rho=0.480$, $p=0.00016$). Hope-waypower and distress show weaker or non-significant external correlations and remain secondary lexical indicators.

\begin{table}[t]
\caption{Selected external lexical comparisons across 57 occurrence-period cells.}\label{tab:S5-validation}
\centering\small
\begin{tabularx}{\linewidth}{>{\raggedright\arraybackslash}p{0.52\linewidth}XX}
\toprule
Comparison & Pearson ($r$, $p$) & Spearman ($\rho$, $p$) \\
\midrule
labMT positive-word share vs happiness prevalence & 0.354, 0.0069 & 0.373, 0.0043 \\
labMT happiness mean vs happiness prevalence & 0.338, 0.0102 & 0.200, 0.1361 \\
LIWC positive emotion vs happiness prevalence & 0.565, $<0.00001$ & 0.578, $<0.00001$ \\
LIWC social process vs collective capability prevalence & 0.356, 0.0065 & 0.480, 0.00016 \\
LIWC future vs hope-waypower prevalence & -0.098, 0.468 & 0.101, 0.457 \\
\bottomrule
\end{tabularx}
\end{table}

\begin{figure*}[t]
\centering
\includegraphics[width=0.88\textwidth]{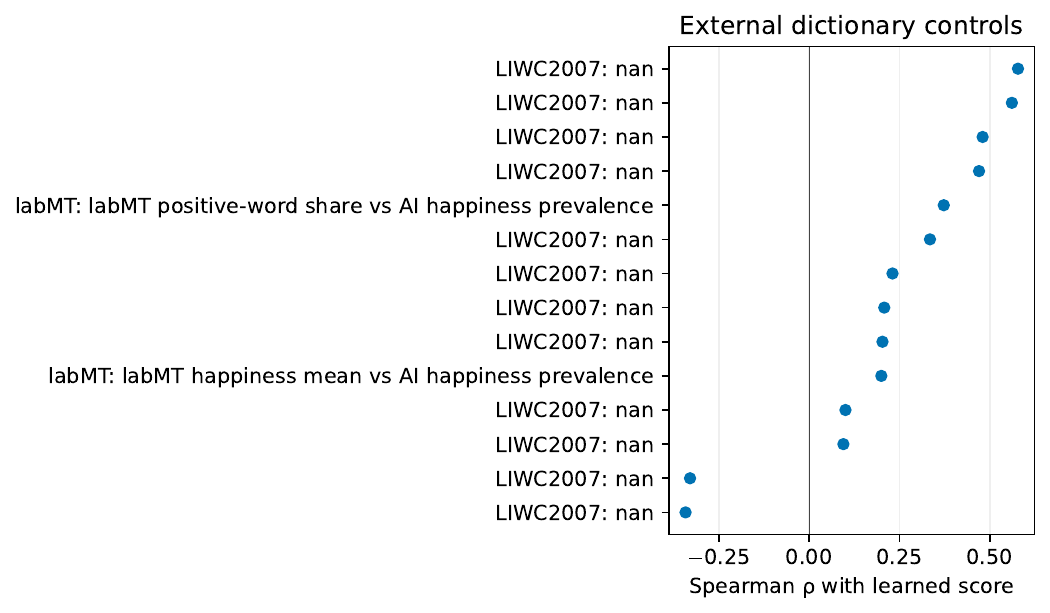}
\caption{External lexical validation of the learned expressed-well-being scores. The comparisons provide language-level convergent checks only.}\label{fig:S2-validation}
\end{figure*}

\subsection{Composition and text-template robustness}\label{subsec:si-robustness}

The direction of the event-period coefficient is preserved under the following pre-specified changes: excluding retweets (+8.62 pp), restricting to English non-retweets (+8.62 pp), author-equal weighting (+9.04 pp), trimming authors above the frozen activity cutoff (+8.45 pp), exact text deduplication (+8.44 pp), and normalized text deduplication (+8.43 pp). Leave-one-campaign and leave-one-occurrence estimates are supplied in the machine-readable SI tables.

\begin{table}[t]
\caption{Event-period happiness estimates under composition and text-template sensitivities.}\label{tab:S6-robustness}
\centering\small
\begin{tabular}{lrr}
\toprule
Specification & Estimate (percentage points) & Direction \\
\midrule
Primary frozen sample & 9.02 & positive \\
Original posts only & 8.62 & positive \\
English original posts only & 8.62 & positive \\
Author-equal daily weighting & 9.04 & positive \\
High-frequency-author trimmed & 8.45 & positive \\
Exact text deduplication & 8.44 & positive \\
Normalized text deduplication & 8.43 & positive \\
\bottomrule
\end{tabular}
\end{table}

\subsection{Independent model-annotation reproducibility audit}\label{subsec:annotation}

Table~S7 summarizes agreement when two independent model coders applied the frozen codebook to the same 451 English posts without seeing the existing scores. This audit concerns reproducibility of model application; it does not provide human inter-rater reliability or criterion validity.

\begin{table}[t]
\caption{Independent model-annotation agreement audit.}\label{tab:S7-annotation}
\centering\small
\begin{tabularx}{\linewidth}{>{\raggedright\arraybackslash}p{0.48\linewidth}X}
\toprule
Construct & Krippendorff's nominal $\alpha$ \\
\midrule
Happiness & 0.811 \\
Distress & 0.519 \\
Hope & 0.321 \\
Collective capability & 0.374 \\
Action & 0.344 \\
\bottomrule
\end{tabularx}
\end{table}

\section{Retweet linkage and cascade audit}\label{sec:network}

The formal retweet table contains 289,532 retweet rows. Exact source matching yields 79,474 matched edges (27.45\% of formal retweet rows), 6,436 unique source posts with matched edges, and 62,317 unique retweeter identifiers. The unresolved-source rows are dominated by scientific-notation precision loss (207,539 rows), so matched cascade statistics are lower bounds. Among matched edges, source and child identifiers are exact and source/child happiness scores are identical in 100\% of cases.

\begin{table}[t]
\caption{Retweet-linkage coverage and cascade estimands.}\label{tab:S8-network}
\centering\small
\begin{tabularx}{\linewidth}{>{\raggedright\arraybackslash}p{0.48\linewidth}X}
\toprule
Quantity & Value \\
\midrule
Formal retweet rows & 289,532 \\
Matched edge rows & 79,474 (27.45\%) \\
Sources in model universe & 68,078 \\
Sources with matched edges & 6,436 (9.45\%) \\
Missing source-ID rows attributable to scientific notation & 207,539 \\
Source/child happiness score exact equality & 100\% \\
\midrule
Hurdle: any observed cascade, OR & 0.457 (95\% CI 0.233--0.896), $p=0.0225$ \\
Hurdle: positive-count IRR & 0.302 (95\% CI 0.047--1.928), $p=0.2057$ \\
First-retweet timing coefficient & 0.089 (95\% CI -0.190--0.368), $p=0.5317$ \\
\bottomrule
\end{tabularx}
\end{table}

Table~S8 provides the complete linkage and cascade audit. Because retweet text duplicates the source, these estimates concern exposure and cascade selection, not recipient emotional contagion. This interpretation follows the distinction between textual diffusion and independent emotional outcomes in social-network research\cite{kramer2014contagion,brady2017emotion,bakshy2015exposure}.

\backmatter

\bibliography{references}